\documentclass[10pt,twocolumn,letterpaper]{article}

\usepackage[pagenumbers]{iccv} % To force page numbers, e.g. for an arXiv version

\usepackage{graphicx}
\graphicspath{{./fig/}}

\definecolor{iccvblue}{rgb}{0.21,0.49,0.74}
\usepackage[pagebackref,breaklinks,colorlinks,allcolors=iccvblue]{hyperref}

\usepackage{breakurl}

\usepackage[accsupp]{axessibility}  % Improves PDF readability for those with disabilities.

\def\paperID{****} % *** Enter the Paper ID here
\def\confName{ICCV}
\def\confYear{2025}

\title{Structure-aware Contrastive Learning for Diagram Understanding of Multimodal Models}

\author{Hiroshi Sasaki\\
The Japan Research Institute, Limited\\
{\tt\small sasaki.hiroshi@jri.co.jp}
}

\begin{document}
\maketitle
\begin{abstract}

Multimodal models, such as the Contrastive Language-Image Pre-training (CLIP) model, have demonstrated remarkable success in aligning visual and linguistic representations. However, these models exhibit limitations when applied to specialised visual domains, such as diagrams, which encode structured, symbolic information distinct from that of natural imagery.

In this paper, we introduce a novel training paradigm explicitly designed to enhance the comprehension of diagrammatic images within vision-language models.
Our approach uses ``hard'' samples for our proposed contrastive learning that incorporates two specialised loss functions that leverage the inherent structural properties of diagrams.
By integrating these objectives into model training, our method enables models to develop a more structured and semantically coherent understanding of diagrammatic content.

We empirically validate our approach on a benchmark dataset of flowcharts, as a representative class of diagrammatic imagery, demonstrating substantial improvements over standard CLIP and conventional hard negative CLIP learning paradigms for both image-text matching and visual question answering tasks. Our findings underscore the significance of tailored training strategies for specialised tasks and contribute to advancing diagrammatic understanding within the broader landscape of vision-language integration.
\end{abstract}
    
\section{Introduction}
\label{sec:intro}

\begin{figure}[t]
    \centering
    \includegraphics[width=0.9\linewidth]{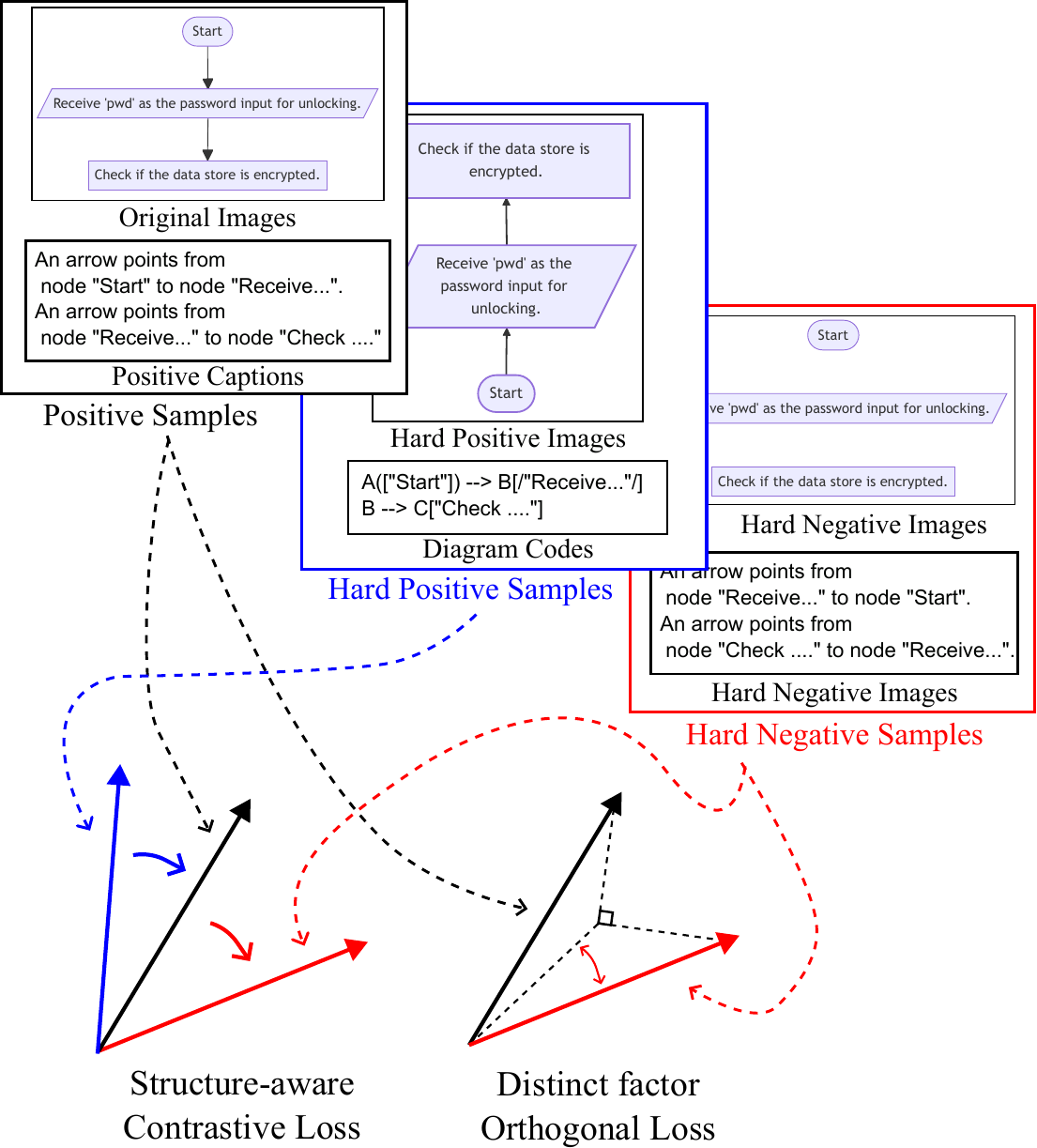}
    \caption{Conceptual illustration of our proposed method. Diagram images and their definition codes are used to generate positive captions, hard positive images, and hard negative pairs. These variants are incorporated into our training framework with two loss functions: the structure-aware contrastive loss, which encourages similarity among positive pairs and dissimilarity between positive-negative pairs, and the distinct factor orthogonal loss, which disentangles shared and uncorrelated factors between correct samples and their negatives.}
    \label{fig:overview}
\end{figure}

The intersection of vision and language has garnered significant attention in recent years, particularly with the rise of powerful multimodal models such as CLIP (Contrastive Language-Image Pre-training)~\cite{radford2021learning}. These models have revolutionised a variety of applications by enabling cross-modal understanding, allowing models to interpret and generate both visual and textual data in a unified framework.
However, despite their impressive performance in handling natural images, current models face significant challenges when applied to specialised visual content such as diagrams, schematics, and technical illustrations. The abstract, structured, and symbolic nature of diagram images introduces complexities that conventional vision-language models (VLMs) struggle to capture effectively.

A primary bottleneck in current multimodal training paradigms is the focus on general visual content, with less emphasis on structured, non-photographic imagery such as diagrams. Whilst CLIP and similar models excel in associating textual descriptions with natural images, they often fail to understand the intricacies of diagrammatic representations, which involve specific relationships between visual elements and textual annotations (e.g., labels, arrows, or graphs). Furthermore, current training methods rely heavily on large-scale paired image-text datasets such as LAION~\cite{schuhmann2022laionb}, which are predominantly focused on natural scenes. As a result, these models lack the nuanced ability to process and relate diagrammatic images with their corresponding textual descriptions, often leading to reduced performance in tasks like diagram interpretation, technical question answering (QA), or knowledge extraction.

In this paper, we propose a novel training methodology designed to enhance the understanding and interpretation of diagram images in VLMs~(\cref{fig:overview}). Inspired by TripletCLIP~\cite{patel2024tripletclip}, we employ contrastive learning with synthetic hard negative samples, introducing a newly designed procedure for hard negative synthesis. Additionally, we propose a training objective that incorporates two specialised loss functions, specifically tailored to leverage the unique structural and semantic features of diagrammatic content.
We introduce a semantic subdivision technique for diagram images and their corresponding textual explanations, enabling the data to be adapted for input into a standard CLIP model in \cref{sec:granulation}. We then apply an editing method to generate hard positive samples—visually distinct yet semantically identical—and hard negative samples—visually similar yet semantically distinct, which is described in \cref{sec:editing}.
Subsequently, we detail our contrastive learning approach, which encourages the model to effectively differentiate between positive and negative samples while disentangling shared representational factors from uncorrelated ones in \cref{sec:saclip}.

In this study, we focus on flowcharts as a representative example of diagrammatic content. Our empirical evaluation highlights the effectiveness of the proposed approach, significantly improving the CLIP model's diagram understanding capabilities, as evidenced by superior performance on a image-text matching task and a visual question answering (VQA) task involving flowchart imagery in \cref{sec:evaluation}.

The primary contributions of this work are as follows:
\begin{itemize}
    \item We introduce a novel preprocessing technique for generating hard positive and negative image-caption pairs within diagrammatic datasets, emphasising subtle yet critical differences that often challenge conventional CLIP models in diagram interpretation.
    \item We propose a novel training objective that incorporates two distinct loss functions, designed to encourage the model to differentiate between semantically valid diagrammatic relationships and their counterfactual counterparts, and facilitate the disentanglement that isolate shared representational factors between correct samples and hard negatives from uncorrelated factors.
    \item Our experimental results validate the efficacy of the proposed approach, demonstrating that it significantly enhances the diagram understanding capabilities of the CLIP model, as evidenced by superior performance gains across multiple tasks including image-text matching and visual question answering within diagram imagery.
\end{itemize}

\section{Related Work}

\paragraph{Contrastive vision-language models}
In recent years, the integration of vision and language models has seen substantial advances, largely driven by the development of large-scale pre-trained models that can learn joint representations of both modalities. ViLBERT~\cite{lu2019vilbert}, VisualBERT~\cite{harold2019visualbert} and UNITER~\cite{chen2020uniter}, which employed task-specific supervision and typically used pre-trained vision encoders (e.g., CNNs or region-based features from Faster R-CNN~\cite{ren2017faster}) along with text encoders (e.g., BERT~\cite{devlin2019bert}), were proposed to learn joint representations of vision and language from large-scale datasets. This model pre-trained both textual and visual components on a variety of tasks, such as image captioning, VQA, and visual reasoning, achieving state-of-the-art performance across several benchmarks.

CLIP~\cite{radford2021learning} has been a landmark model in this area, showcasing impressive capabilities in a wide variety of vision-language tasks using a contrastive learning approach.
Contrastive learning is rooted in the idea of learning representations by contrasting positive pairs (similar examples) and negative pairs (dissimilar examples). This technique has been widely used in unsupervised learning, where the goal is to learn meaningful representations without relying on explicit labels. Early successes of contrastive learning were seen in the field of computer vision, with methods like SimCLR~\cite{chen2020simple} and MoCo~\cite{he2020momentum}, which applied contrastive loss to image data by contrasting different augmentations of the same image (positive pair) against different images (negative pairs).
CLIP’s use of contrastive learning has gained significant attention for its simplicity and effectiveness in bridging the gap between vision and language. The contrastive loss function maximises the cosine similarity between the representations of matching image-text pairs and minimises the similarity between non-matching pairs. 

Whilst CLIP has achieved notable success across a wide array of vision-language tasks, it tends to focus primarily on a limited subset of the input~\cite{paiss2022no}, often centering on nouns (i.e., individual objects), while neglecting other critical elements such as adjectives and attributes. In this work, we adapt the CLIP training paradigm to place greater emphasis on non-object-related information, which plays a crucial role in the interpretation of diagrams, such as the relationships between nodes and other structural components.

\paragraph{Hard negative contrastive learning}
One critical aspect of contrastive learning is the choice of negative samples, which play a pivotal role in the quality of the learnt representations. The use of hard negative samples—those that are visually or semantically similar to the positive sample but not identical—has become a key strategy to enhance the learning process. The motivation behind hard negative sampling is rooted in the idea that the model will benefit more from distinguishing between subtle, difficult-to-differentiate pairs than from easy negatives that are far apart in the feature space. By including such hard negatives, the model can learn finer-grained distinctions between instances, leading to representations that generalise better to a wide range of tasks.

VSE++~\cite{faghri2018vse++}, NegCLIP and Triplet CLIP~\cite{patel2024tripletclip} incorporate hard negatives into contrastive learning for VLMs. In the triplet-based methods, the loss function operates on triplets of samples, consisting of an anchor, a positive and negative samples from a different modality. Hard negatives in this context refer to those negative samples that are closest to the anchor in the embedding space but still belong to a different class. The inclusion of hard negatives in triplet loss-based models has been shown to enhance the discriminative power of the learnt representations and improve performance on tasks such as face recognition and retrieval.
Hard negatives have also been employed to train CLIP models specialised in object counting by generating synthetic counterfactual captions that alter the object counting information in the text for triplet contrastive learning~\cite{paiss2023teaching}.

\begin{figure}
    \centering
    \begin{subfigure}{0.3\linewidth}
      \includegraphics[width=1.0\linewidth]{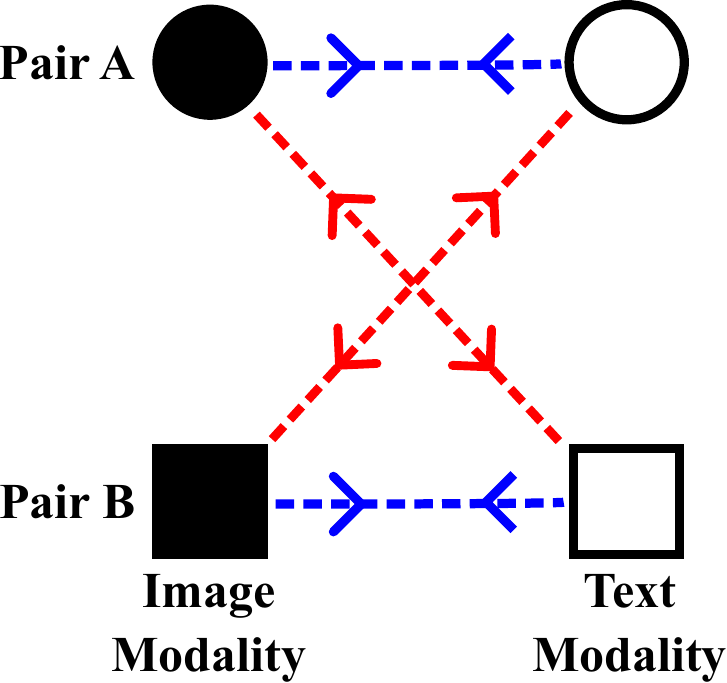}
      \caption{CLIP}
      \label{fig:clip}
    \end{subfigure}
    \hspace{9mm}
    \begin{subfigure}{0.37\linewidth}
      \includegraphics[width=1.0\linewidth]{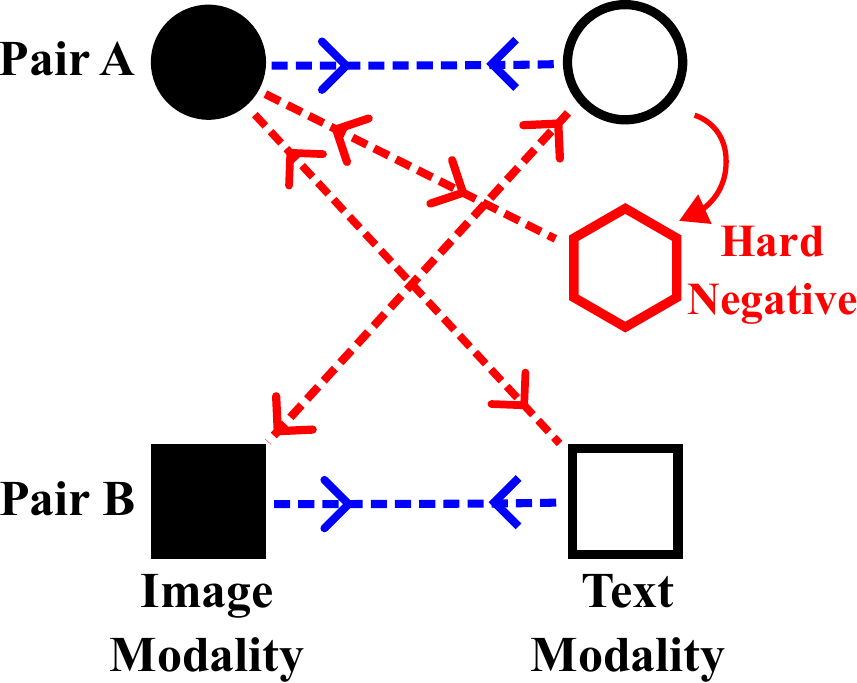}
      \caption{NegCLIP~\cite{patel2024tripletclip}}
      \label{fig:negclip}
    \end{subfigure}
    \begin{subfigure}{0.36\linewidth}
      \includegraphics[width=1.0\linewidth]{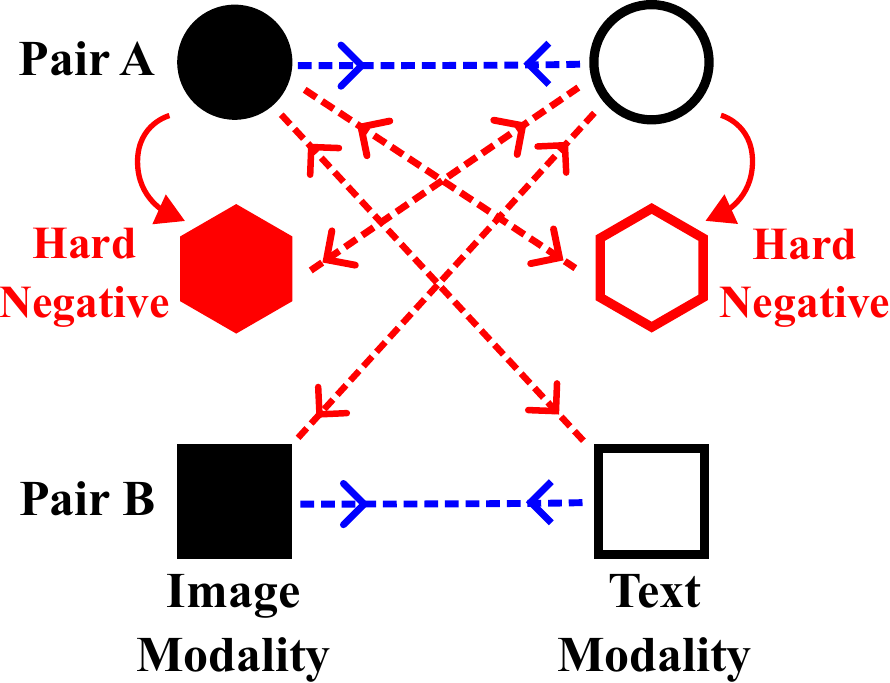}
      \caption{TripletCLIP~\cite{patel2024tripletclip}}
      \label{fig:tripletclip}
    \end{subfigure}
    \hspace{5mm}
    \begin{subfigure}{0.36\linewidth}
      \vspace{3mm}
      \includegraphics[width=1.0\linewidth]{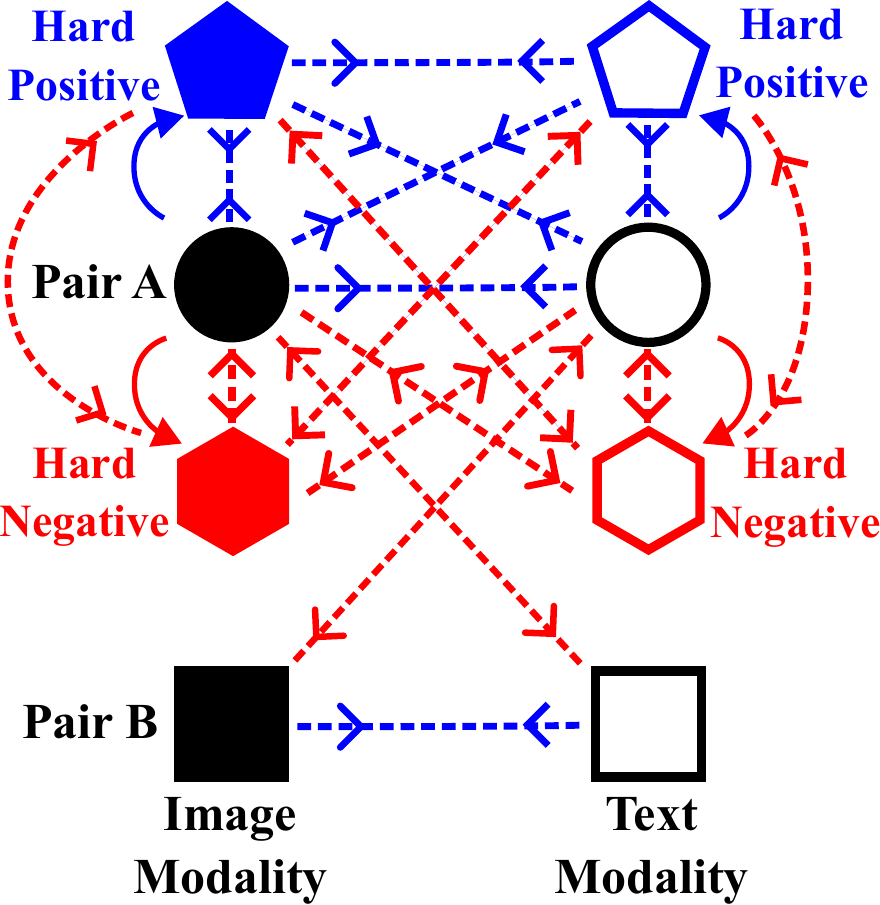}
      \caption{Ours}
      \label{fig:saclip}
    \end{subfigure}
    \caption{
        Conceptual illustration of contrastive learning strategies in CLIP~\cite{radford2021learning}, NegCLIP, TripletCLIP~\cite{patel2024tripletclip}, and our proposed method. (a) CLIP minimises the distance between matched inter-modal pairs while maximising the distance between mismatched ones. (b–-c) NegCLIP and TripletCLIP utilise hard negative samples, encouraging the model to further increase the distance between the anchor and its semantically similar but incorrect counterparts in a different modality. (d) Our method additionally generates hard positive samples and minimises distances across both inter-modal and intra-modal positive pairs, while maximising those across both types of negative pairs.
        }
    \label{fig:contrastive_learning}
  \end{figure}

Inspired by~\cite{patel2024tripletclip}~\cite{paiss2023teaching}, our method employ the synthetic hard negative strategy and hard negative contrastive learning through the manipulation of nodes and relational connectors (e.g., arrows) in diagrams. However, we extend this framework with additional strategies aimed at enhancing performance. In addition to the creation of hard negatives, we also generate hard positive samples—visually distinct but semantically identical counterparts—to enable the model to learn subtle yet critical distinctions within diagrammatic content. These curated hard positive and negative instances are utilised not only for inter-modal triplet construction but also for intra-modal manipulations (illustrated in \cref{fig:contrastive_learning} and discussed further in \cref{sec:method}). Moreover, our training methodology not only leverages contrastive learning but also accounts for the shared information between positive and negative samples, ensuring that the model retains its ability to capture coexistent information without losing crucial representational nuances.

\paragraph{Diagram specific vision models}
Whilst general-purpose VLMs like CLIP have demonstrated remarkable success across a variety of multimodal tasks, specialised models have been developed to address the unique challenges of diagram understanding.

A prominent approach within this domain involves finetuning existing models on diagram-specific datasets~\cite{masry2023unichart}~\cite{zhang2025mavis}. UniChart~\cite{masry2023unichart} is a pretrained model tailored for chart comprehension and reasoning, which utilises the Donut architecture~\cite{kim2022ocr} and is finetuned with a chart image dataset. MAVIS~\cite{zhang2025mavis} proposes a mathematical visual instruction tuning framework for VLMs by employing a math-specific vision encoder called CLIP-Math, which is a CLIP model finetuned with mathematical diagram images to enable visual mathematical problem-solving.

Our approach aligns with the finetuning paradigm, but introduces modifications to the training framework by incorporating our specialised contrastive learning and disentanglement learning, which are designed to further enhance the model's ability to comprehend the structural aspects of diagrammatic imagery.

\section{Method}
\label{sec:method}
Our approach assumes that the training dataset consists of diagram images and their associated textual descriptions, which define the underlying structure of the diagram (e.g., Mermaid code~\cite{std:mermaid}). These textual representations are referred to as "diagram codes" or simply "codes" throughout this paper.
Given that a standard CLIP model typically supports limited image and text sizes, which may be insufficient to represent the full complexity of each diagram in the dataset, we propose a method to decompose and reconstruct each diagram into a set of simpler subparts, a process we refer to as "granulation" (detailed in \cref{sec:granulation}). 
These granulated subparts are subsequently manipulated to create hard positive and negative samples, as outlined in  \cref{sec:editing}. 
These hard samples are then utilised in our proposed structure-aware contrastive learning framework, which is described in \cref{sec:saclip}. A visual summary of the process is provided in \cref{fig:overview}.

\subsection{Diagrammic data granulation}
\label{sec:granulation}
\begin{figure}[t]
  \centering
   \includegraphics[width=1.0\linewidth]{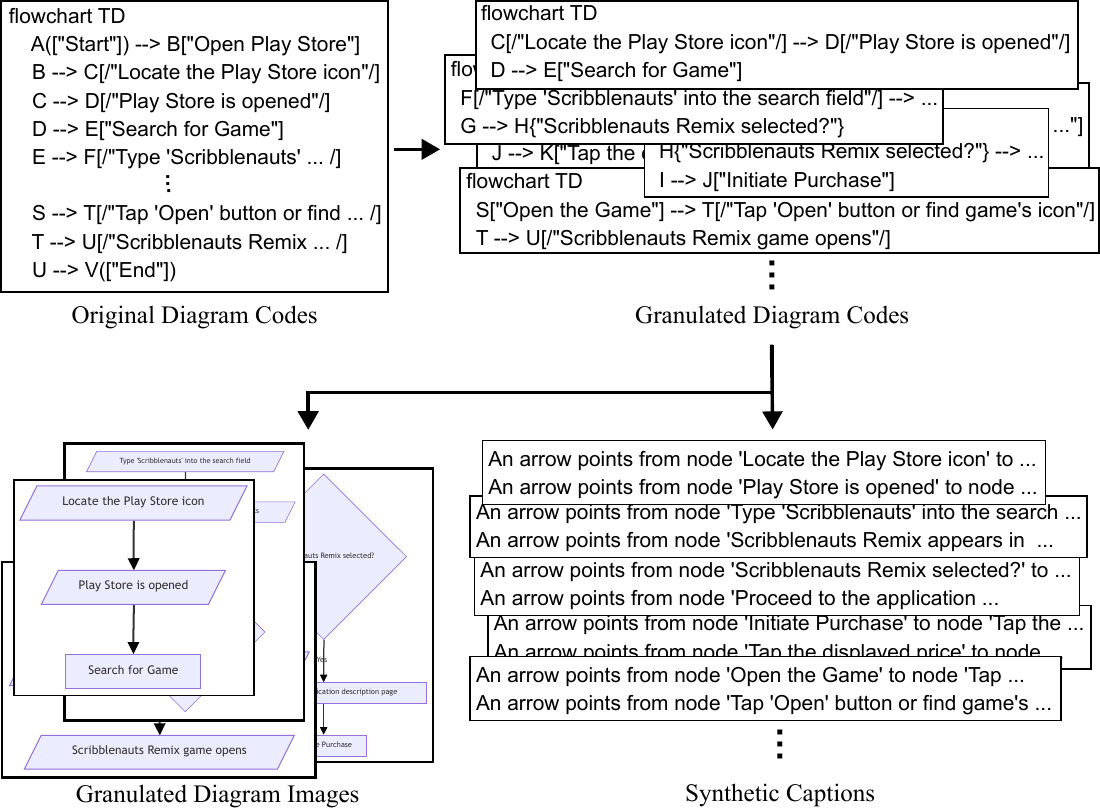}
   \caption{The diagrammatic data granulation process flow. The original diagram codes are decomposed and subsequently reconstructed into a series of smaller, modular diagram codes, which are then transformed into image files and text captions.}
   \label{fig:granulation}
   \vspace{-1mm}
\end{figure}

The granulation process, as illustrated in \cref{fig:granulation}, involves extracting all combinations of adjacent triplets of nodes from each diagram code. These triplets are used to regenerate simplified versions of the original codes. The resulting granulated codes are aligned their flows top-to-down and then converted into diagram images, both as raster graphics for model input and as vector graphics for the hard sample generation step in \cref{sec:editing}.
Additionally, the granulated codes are employed to synthesise corresponding textual descriptions, following the template: {\it ``An arrow points from node $\langle A \rangle$ to node $\langle B \rangle$''}, where $\langle A \rangle$ and $\langle B \rangle$ represent the names of the nodes from and to which the arrow is directed, respectively. These descriptions are used as captions during the hard sample generation phase (\cref{sec:editing}).
 
\subsection{Diagrammatic hard positive / negative data synthesis}
\label{sec:editing}
\begin{figure}[t]
    \centering
     \includegraphics[width=1.0\linewidth]{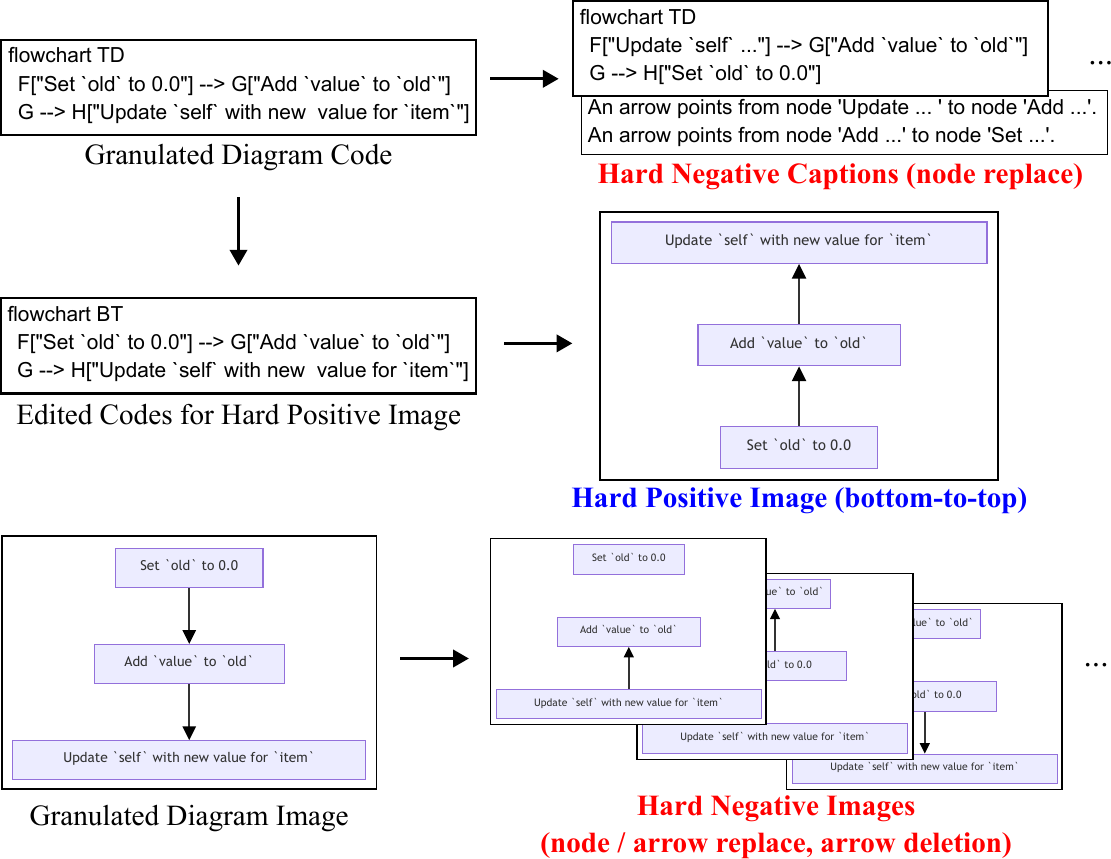}
     \caption{The synthesis process of hard positive and hard negative samples. Hard positive samples are the diagram codes along with the diagram images altered its visual orientation of the flow. Conversely, hard negative samples are constructed by permuting the nodes and edges of the diagram codes, captions and images.}
     \label{fig:hard_sample}
  \end{figure}

Let $\mathcal{V}$ and $\mathcal{T}$ represent image and text modalities, respectively, and $D=\{d_i\}^{N^D}_{i=1}=\{(v_i,t_i,c_i)\}^{N^D}_{i=1}$ denotes the training dataset, where $v_i \in \mathcal{V}$ is the granulated diagram images and $t_i,c_i \in \mathcal{T}$ are the corresponding description and codes generated in \cref{sec:granulation}.
We define the generation of hard positive and negative samples as follows:
\begin{eqnarray}
    V^p_i=\{v^p_{i,j}\}^{N^\textnormal{VP}}_{j=1} = P^v(v_i,c_i), \\
    V^n_i=\{v^n_{i,j}\}^{N^\textnormal{VN}}_{j=1} = N^v(v_i,c_i), \\
    T^p_i=\{t^p_{i,j}\}^{N^\textnormal{TP}}_{j=1} = P^t(t_i,c_i), \\
    T^n_i=\{t^n_{i,j}\}^{N^\textnormal{TN}}_{j=1} = N^t(t_i,c_i),
    \label{eqn:edit}
\end{eqnarray}
where $P^v, N^v, P^t, N^t$ represent the functions that generate hard positive images, hard negative images, hard positive captions, and hard negative captions, respectively. These functions are governed by a set of editing rules outlined below:
\begin{description}
    \item[Rule for Hard Positive Images ($P^v$)] \hfill \\
    Reverse the flow direction from top-down to bottom-up (since all original diagrams are defined with a top-down flow). 
    \item[Rule for Hard Negative Images ($N^v$)] \hfill \\
    Apply a combination of the following perturbations:
    \begin{itemize}
        \item Randomly swap the labels of selected nodes.
        \item Reverse the direction of randomly selected arrows.
        \item Remove a subset of arrows at random.
        \item Apply the above operations after $P^v$.
    \end{itemize} 
    \item[Rule for Hard Positive Captions ($P^t$)] \hfill \\
    The code of the diagram ($=c_i$). 
    \item[Rule for Hard Negative Captions ($N^t$)] \hfill \\
    Apply a combination of the following semantic distortions:
    \begin{itemize}
        \item Randomly swap the labels of selected nodes in the natural language description. \hfill 
        \item Randomly swap the labels of nodes within $c_i$.
    \end{itemize} 
\end{description}
These hard positive and negative samples are generated by applying the respective editing rules, as visually illustrated in \cref{fig:hard_sample}.

\subsection{Structure-aware contrastive learning}
\label{sec:saclip}
Our proposed training procedure, referred to as structure-aware contrastive learning, extends the standard CLIP training framework. The contrastive loss function of the standard CLIP is defined using the InfoNCE loss~\cite{oord2018representation}, which is expressed as:
\begin{align}
    \mathcal{L}_\textnormal{CL} = \frac{1}{2} \mathbb{E}_{v_i, t_i}\left[ -\log \frac{\exp \langle F_v(v_i), F_t(t_i) \rangle}{\mathbb{E}_{v_i, t_{j\neq i}}[\langle F_v(v_i), F_t(t_j) \rangle]} \right. \nonumber \\
    \left. - \log \frac{\exp \langle F_v(v_i), F_t(t_i) \rangle}{\mathbb{E}_{v_{j\neq i}, t_i}[\langle F_v(v_j), F_t(t_i) \rangle]}\right] ,
    \label{eqn:clip_loss}
\end{align}
where $\langle \cdot, \cdot \rangle$ denotes cosine similarity, $F_v$ and $F_t$ represent the vision and text encoder, respectively.
For clarity, feature normalization and the temperature parameter are omitted in the equations in this paper.
To enable the CLIP model to effectively distinguish between the hard negative samples (generated in \cref{sec:editing}) and positive samples (which include both the original and hard positive samples from \cref{sec:editing}), we propose two novel loss functions: structure-aware contrastive loss and distinct factor orthogonal loss.

\paragraph{Structure-aware contrastive loss}
Our approach aims to simultaneously draw hard positive samples closer to the original anchors while pushing hard negative samples further away. To achieve this, we propose the Structure-aware Contrastive (SC) Loss, which extends NegCLIP and Triplet losses~\cite{patel2024tripletclip} by incorporating both intra-modal and inter-modal distances of these samples. Unlike prior methods that consider only inter-modal distances between original and hard negatives, the SC loss accounts for all pairwise relations among original, hard positive, and hard negative samples. This encourages more coherent local structure and better cross-modal alignment, as illustrated in \cref{fig:contrastive_learning}.

The SC loss is calculated using the positive similarity $\mathcal{S}^p(d_i)$ and negative similarity $\mathcal{S}^n(d_i)$ between samples, which are defined as follows:
\begin{align}
    \mathcal{S}^p(d_i) &= \nonumber \\
    &\mathbb{E}_{v^p\sim V^p_i, t^p\sim T^p_i} [\exp \langle F_v(v_i), F_v(v^p) \rangle + \nonumber \\
    &~~\exp\langle F_t(t_i), F_t(t^p) \rangle + \exp \langle F_v(v_i), F_t(t^p) \rangle + \nonumber \\
    &~~\exp \langle F_t(t_i), F_v(v^p) \rangle ],
    \label{eqn:positive_similarity}
\end{align}
\begin{align}
    \mathcal{S}^n(d_i) &= \nonumber \\
    &\mathbb{E}_{v^n\sim V^n_i, t^n\sim T^n_i} [\exp \langle F_v(v_i), F_v(v^n) \rangle + \nonumber \\
    &~~\exp \langle F_t(t_i), F_t(t^n) \rangle + \exp \langle F_v(v_i), F_t(t^n) \rangle + \nonumber \\
    &~~\exp \langle F_t(t_i), F_v(v^n) \rangle].
    \label{eqn:negative_similarity}
\end{align}
The SC loss is then formulated as:
\begin{align}
    \mathcal{L}_\textnormal{SC} = \mathbb{E}_{d_i \sim D} \left[ -\log \frac{\mathcal{S}^p(d_i)}{\mathcal{S}^p(d_i)+\mathcal{S}^n(d_i)} \right].
    \label{eqn:rcloss}
\end{align}
This loss encourages the model to minimise the distance between the original and hard positive samples, while maximising the distance between the original and hard negative samples.

\paragraph{Distinct factor orthogonal loss}
Whilst the SC loss effectively encourages discrimination between the original and hard negative samples, it does not account for the fact that both the original and hard negative samples may share important semantic information, such as node names. To prevent the SC loss from disrupting these shared components of the embedding vectors, we introduce an additional regularisation term, the Distinct factor Orthogonal (DO) loss, to preserve such shared information.

We assume that the shared and distinct information in the embedding space can be represented as linear combinations of vectors, denoted by:
\begin{align}
    &F_*(*_i) = z_{*_i} = z^s_{*_i} + z'_{*_i}, \\
    &F_*(*^n_{i,j}) = z_{*^n_{i,j}} = z^s_{*_i} + z'_{*^n_{i,j}}, \\
    &(*=v|t:\textnormal{vision or text modarity}) \nonumber
    \label{eqn:shared_vectors}
\end{align}
where $z^s_*$ and $z'_*$ represent the shared and distinct vectors, respectively.
Our goal is to make the distinct factors $z'_{*_i}$ and $\{z'_{*^n_{i,j}}\}_{j=1,\dots}$ as orthogonal as possible, thereby ensuring that the shared information is preserved while the distinct components are decoupled. Whilst the exact angles between these vectors cannot be directly computed due to the unknown nature of the shared and distinct components, we leverage Thales's theorem to approximate their orthogonality, assuming their embedding space is locally approximated to Euclidean space.
\begin{figure}[t]
    \centering
    \includegraphics[width=0.6\linewidth]{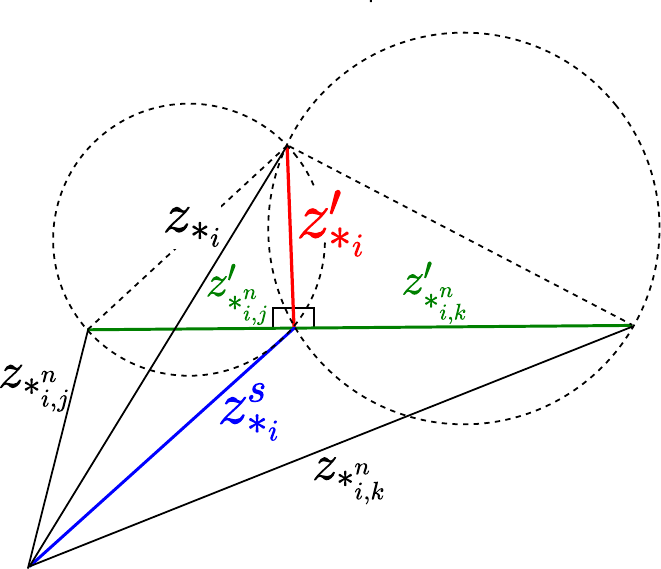}
    \caption{The visualisation of the disentanglement in the original and hard negative samples, framed within the context of Thales's theorem. The shared factor vector $z^s_*$ is positioned on the intersection of the circles whose diameters are defined by the line segments connecting the vectors representing the original ($z_{*_i}$) and hard negative samples ($z_{*^n_{i,j}}, z_{*^n_{i,k}}$).}
    \label{fig:shared_vector}
\end{figure}
According to Thales's theorem, the position of the shared vector $z^s_*$ lies on the intersection of the circles whose diameters are the lines between the distinct vectors $z'_{*_i}$ and $\{z'_{*^n_{i,j}}\}_{j=1,\dots}$, provided that these vectors are orthogonal (\cref{fig:shared_vector}). Thus, we can express this relationship as:
\begin{equation}
    \left| z^s_{*_i} - \frac{1}{2} (z_{*_i} + z_{*^n_{i,j}}) \right| = \left| \frac{1}{2} (z_{*_i} - z_{*^n_{i,j}}) \right|.
    \label{eqn:thales}
\end{equation}
Using the relation \cref{eqn:thales}, $z^s_{*_i}$ can be found by solving the following equation:
\begin{align}
    Z_{*_i} (z^s_{*_i})^T = z^c_{*_i},
    \label{eqn:z}
\end{align}
where
\begin{align}
    Z_{*_i} = \left[ (z_{*^n_{i,2}}-z_{*^n_{i,1}})~~(z_{*^n_{i,3}}-z_{*^n_{i,2}})~~\cdots \right]^T, \nonumber \\
    z^c_{*_i} = 
    \begin{bmatrix}
        r^T_2r_2-r^T_1r_1+c^T_2c_2-c^T_1c_1  \\
        r^T_3r_3-r^T_2r_2+c^T_3c_3-c^T_2c_2  \\
        \vdots
    \end{bmatrix}, \nonumber \\
    r_i = \left| \frac{1}{2} (z_{*_i} - z_{*^n_{i,j}}) \right|,~~
    c_i = \left| \frac{1}{2} (z_{*_i} + z_{*^n_{i,j}}) \right|. \nonumber
\end{align}
We can then approximately solve for $z^s_{*_i}$ by applying the Moore–Penrose inverse to compute the optimal solution $\hat{z}^s_{*_i}$ as:
\begin{align}
    \hat{z}^s_{*_i} = \left( Z^+_{*_i} z^c_{*_i} \right)^T,
    \label{eqn:z}
\end{align}
where $Z^+_{*_i}$ is the Moore–Penrose inverse of $Z_{*_i}$.
The DO loss is finally defined as:
\begin{align}
    \mathcal{L}_\textnormal{DO} = 
    \mathbb{E}_{(v_i,t_i) \sim D} \left[ \frac{1}{2}\left| Z_{v_i} (\hat{z}^s_{v_i})^T - z^c_{v_i} \right| \right.\nonumber \\
    \left. +\frac{1}{2}\left| Z_{t_i} (\hat{z}^s_{t_i})^T - z^c_{t_i} \right| \right].
    \label{eqn:io_loss}
\end{align}

\paragraph{Total loss}
The total loss for our method combines both the SC and DO losses:
\begin{align}
    \mathcal{L} = \mathcal{L}_\textnormal{CL}+\lambda_\textnormal{SC}\mathcal{L}_\textnormal{SC}+\lambda_\textnormal{DO}\mathcal{L}_\textnormal{DO},
    \label{eqn:total_loss}
\end{align}
where $\lambda_\textnormal{SC}$ and $\lambda_\textnormal{DO}$ are hyperparameters controlling the relative contributions of the SC and DO losses.
By optimising the vision and text encoders using the total loss function in \cref{eqn:total_loss}, the model is trained to effectively capture the subtle structural relationships within diagrammatic images while preserving shared information in the embedding space.

\section{Evaluation}
\label{sec:evaluation}
We evaluate the proposed method in \cref{sec:method} on flowcharts, a quintessential example of diagrams.

\subsection{Dataset Preparation}
\label{sec:dataset}
For the experiments, we utilise the publicly available FlowVQA dataset~\cite{singh2024flowvqa}, which includes a set of flowchart images, corresponding textual descriptions, Mermaid code representations, and QA pairs. The Mermaid codes are parsed and transformed according to the procedures in \cref{sec:granulation} and \cref{sec:editing}, serving as input for the CLIP finetuning described in \cref{sec:clip_training}.

The granulation process involves parsing the Mermaid code, where the flow direction is defined as top-to-bottom. This directional definition is preserved during the granulation step. The resulting flowchart images are rendered as SVG~\cite{std:svg} files using the mermaid-cli~\cite{code:mermaid-cli} to be edit for hard negative sample synthesis.

\vspace{-1mm}
\paragraph{Hard sample synthesis}
Following~\cref{sec:editing}, the granulated Mermaid codes are utilised as hard positive captions, while hard positive images are synthesised by inverting the original flow direction (from top-to-bottom to bottom-to-top).
For hard negative sample generation, both captions and images are constructed via the editing operations in~\cref{sec:editing}. During training, these edits are applied dynamically to samples within each batch, ensuring continual exposure to challenging counterexamples and encouraging robustness in the learnt representations.
To ensure consistency and mitigate stochastic variability during evaluation, we precompute a fixed set of six hard negative captions and eight hard negative images for each sample in the test set. These pre-generated negatives are held constant across all evaluation runs, enabling fair and reproducible comparisons across different model variants. 

\begin{table*}[tb]
    \begin{center}
        \caption{Evaluation results for the image-text matching task, reporting Recall@1 (R@1), Recall@5 (R@5), Recall@10 (R@10), and Mean Reciprocal Rank (MRR) across different finetuning strategies.}
        \begin{tabular}{l c c c c c c c c} \hline
                    & \multicolumn{4}{c}{Query: images, Retrieve: captions} & \multicolumn{4}{c}{Query: captions, Retrieve: images} \\
                   & R@1 & R@5 & R@10 & MRR & R@1 & R@5 & R@10 & MRR \\ \hline
            w/o FT~\cite{clip:vit-large-patch14} & 0.57475 & {\bf 0.90347} & {\bf 0.93825 } & 0.71826 & 0.57317 & {\bf 0.89448 } & {\bf 0.93775 } & 0.71054 \\
            CLIP~\cite{radford2021learning} & 0.60913 & 0.78243 & 0.82067 & 0.68605 & 0.63541 & 0.80002 & 0.83697 & 0.70831 \\
            NegCLIP~\cite{patel2024tripletclip} & 0.60429 & 0.72206 & 0.74617 & 0.65737 & 0.63097 & 0.75951 & 0.78658 & 0.68795 \\
            TripletCLIP~\cite{patel2024tripletclip} & 0.66239 & 0.80229 & 0.82818 & 0.72432 & 0.72019 & 0.85762 & 0.88410 & 0.78185 \\ \hline
            SaCLIP (ours) &&&&&&&& \\
            \begin{tabular}{c}$\lambda_\textnormal{SC}=0.1, \lambda_\textnormal{DO}=0.0$\end{tabular} & {\bf 0.74667} & 0.87462 & 0.89685 & {\bf 0.80391} & {\bf 0.76583} & 0.88845 & 0.91088 & {\bf 0.82143} \\
            \begin{tabular}{c}$\lambda_\textnormal{SC}=0.1, \lambda_\textnormal{DO}=0.01$\end{tabular} & 0.72760 & 0.86207 & 0.88707 & 0.78828 & 0.74341 & 0.87630 & 0.89873 & 0.80281 \\
            \begin{tabular}{c}$\lambda_\textnormal{SC}=0.1, \lambda_\textnormal{DO}=0.1$\end{tabular} & 0.72424 & 0.86009 & 0.88756 & 0.78534 & 0.74627 & 0.87738 & 0.89803 & 0.80537 \\ \hline
        \end{tabular}
        \label{tab:ranking}
    \end{center}
\end{table*}

\begin{table*}[tb]
    \begin{center}
        \caption{Evaluation results for the image-text matching task in the presence of hard negative samples, reported using Recall@1 (R@1), Recall@3 (R@3), and Mean Reciprocal Rank (MRR) across different finetuning strategies.}
        \begin{tabular}{l c c c c c c} \hline
                    & \multicolumn{3}{c}{Query: images, Retrieve: captions} & \multicolumn{3}{c}{Query: captions, Retrieve: images} \\
                   & R@1 & R@3 & MRR & R@1 & R@3 & MRR \\ \hline
                   w/o FT~\cite{clip:vit-large-patch14} & 0.212034 & 0.212034 & 0.212034 & 0.439976 & 0.729770 & 0.592791 \\
                   CLIP~\cite{radford2021learning} & 0.497777 & 0.497777 & 0.497777 & 0.750716 & 0.939235 & 0.844134 \\
                   NegCLIP~\cite{patel2024tripletclip} & 0.515957 & 0.515957 & 0.515957 & 0.813556 & 0.970655 & 0.890866 \\
                   TripletCLIP~\cite{patel2024tripletclip} & 0.544907 & 0.544907 & 0.544907 & 0.971940 & 0.994665 & 0.983083 \\ \hline
                   SaCLIP (ours) &&&&&& \\
                   \begin{tabular}{c}$\lambda_\textnormal{SC}=0.1, \lambda_\textnormal{DO}=0.0$\end{tabular} & {\bf 0.664361} & {\bf 0.664361} & {\bf 0.664361} & 0.967395 & {\bf 0.998518} & 0.982573 \\
                   \begin{tabular}{c}$\lambda_\textnormal{SC}=0.1, \lambda_\textnormal{DO}=0.01$\end{tabular} & 0.639166 & 0.639166 & 0.639166 & 0.974706 & 0.998024 & 0.986125 \\
                   \begin{tabular}{c}$\lambda_\textnormal{SC}=0.1, \lambda_\textnormal{DO}=0.1$\end{tabular} & 0.654382 & 0.654382 & 0.654382 & {\bf 0.975002} & 0.998320 & {\bf 0.986421} \\ \hline
        \end{tabular}
        \label{tab:ranking_hard}
    \end{center}
\end{table*}

\subsection{Baselines}
\label{sec:baselines}
We compare our method against the following finetuning strategies:
\begin{description}
    \item[w/o FT] The publicly available CLIP ViT-L/14@336px model~\cite{clip:vit-large-patch14} without any task-specific finetuning (zero-shot baseline).
    \vspace{-2mm}
    \item[CLIP] Finetuned on the flowchart dataset using the standard CLIP loss~\cite{radford2021learning}, without hard negatives.
    \vspace{-2mm}
    \item[NegCLIP] Finetuned with the NegCLIP loss~\cite{patel2024tripletclip}, incorporating hard negative captions.
    \vspace{-2mm}
    \item[TripletCLIP] Finetuned using the TripletCLIP loss~\cite{patel2024tripletclip}, which leverages hard negative images and captions.
    \vspace{-2mm}
    \item[SaCLIP] Our method (see~\cref{sec:method}), which incorporates both hard positive and hard negative image-caption pairs.
\end{description}

\subsection{CLIP Training}
\label{sec:clip_training}
The finetuning in \cref{sec:baselines} is performed using the dataset introduced in \cref{sec:dataset} and the LoRA method~\cite{hu2022lora} applied to all linear layers, with the parameters $\alpha=256$ and $r=64$. Prior to model input, all images are converted to raster format. The training procedure spans 3 epochs with a learning rate of 0.0004, initiated after 8 steps of warmup. For our proposed method, we fix $\lambda_\textnormal{SC}=0.1$ and evaluate the influence of the DO loss by varying $\lambda_\textnormal{DO}=\{ 0, 0.01, 0.1 \}$.

\subsection{Image-text Matching Task Evaluation}
All images and captions from the test set described in~\cref{sec:dataset} are encoded into feature representations using the respective encoders of the models finetuned as in~\cref{sec:baselines}. For each image, the top 10 most similar captions are retrieved, and vice versa, based on cosine similarity. To evaluate image-text matching accuracy, we compute Recall@1 (R@1), Recall@5 (R@5), Recall@10 (R@10), and Mean Reciprocal Rank (MRR) over the retrieved sets. The results are presented in~\cref{tab:ranking}.

All finetuning methods yield notable improvements in R@1, with our proposed method achieving the highest gains. Whilst a general degradation in R@5 and R@10 is observed for some approaches, our method exhibits relatively strong robustness against this trend, resulting in the highest overall improvement in MRR. An ablation study on the DO loss shows limited impact on performance in this experiment; however, its effectiveness becomes more evident in subsequent evaluations.

To further assess robustness, we conduct the same experiment under more challenging conditions involving hard negatives. Specifically, for each image, the top 3 most similar captions are retrieved from a candidate set comprising the correct caption and hard negatives; the same process is applied for each caption. We report R@1, R@3, and MRR in~\cref{tab:ranking_hard}. The results demonstrate that all finetuning strategies enhance performance in this harder setting, with our method again outperforming the others. Notably, the DO loss contributes positively in this scenario, improving the model’s ability to retrieve semantically aligned images despite the presence of hard distractors.

\subsection{VQA Task Evaluation}
We evaluate the impact of our specialised finetuned CLIP model on diagram comprehension within a VQA framework, specifically when integrated into a CLIP-based large language model (LLM). For this purpose, we employ the LLaVA-v1.6-Mistral-7B model~\cite{llava:v1.6-mistral-7b}, a widely adopted publicly available variant of LLaVA~\cite{liu2024visual} that incorporates the CLIP ViT-L/14@336px vision encoder—architecturally identical to the CLIP model used in our experiments. This compatibility makes it a suitable testbed for assessing the effectiveness of our finetuning approach.

Unlike previous experiments that utilise the granulated version of the dataset, this evaluation leverages the original FlowVQA dataset, as the LLM supports higher-resolution inputs via a split-and-merge strategy~\cite{liu2024improved}. To examine the influence of vision encoder quality, we compare model performance when the default encoder is replaced with each of the variants introduced in~\cref{sec:baselines}.

\paragraph{LLaVA Training}
All modules, except for the vision encoders, are finetuned using the LoRA method applied to all linear layers with $\alpha=32$ and $r=16$. The training procedure spans 3 epochs with a learning rate of 0.0004, initiated after 8 steps of warmup.

\paragraph{VQA Performance}
We evaluate each model's performance on the FlowVQA test set by computing the average BERTScore~\cite{zhang2020bertscore} between the model outputs and the corresponding ground truth answers. The results, summarised in \cref{tab:llm_results}, demonstrate that finetuning the vision encoder significantly improves performance, with our method yielding superior precision and F1 score compared to the baseline CLIP finetuning approaches. Notably, the inclusion of the DO loss in our approach contributes to a marked improvement in F1 performance whithin the VQA framework, highlighting its effectiveness in enhancing semantic alignment for diagram comprehension.  

\begin{table}[tb]
    \begin{center}
        \caption{The comparison of the BERTScore~\cite{zhang2020bertscore} performance across different finetuning strategies.}
        \begin{tabular}{l c c c} \hline
                   & Precision & Recall & F1 \\ \hline
                w/o FT~\cite{clip:vit-large-patch14} & 0.4834611 & 0.4259244 & 0.4947443 \\
                CLIP~\cite{radford2021learning} & 0.6891659 & 0.5501751 & 0.6217804 \\
                NegCLIP~\cite{patel2024tripletclip} & 0.6767507 & 0.5554876 & 0.6177384 \\
                TripletCLIP~\cite{patel2024tripletclip} & 0.6874915 &	{\bf 0.5727245} &	0.6284228 \\
                SaCLIP (ours) &&& \\
                \begin{tabular}{c}$\lambda_\textnormal{DO}=0.0$\end{tabular} & {\bf 0.6972192} & 0.5440854 & 0.6146211 \\
                \begin{tabular}{c}$\lambda_\textnormal{DO}=0.01$\end{tabular} & 0.6811157 &	0.5597942 &	0.6242264 \\
                \begin{tabular}{c}$\lambda_\textnormal{DO}=0.1$\end{tabular} & 0.6882976 & 0.5666072 &	{\bf 0.6340190} \\ \hline
        \end{tabular}
        \label{tab:llm_results}
    \end{center}
\end{table}

\subsection{Limitation} 
Whilst the experiments presented in this paper demonstrate the effectiveness of our method in improving performance on flowchart image-text matching and VQA tasks, there are several notable limitations. Firstly, our approach assumes the availability of diagram-specific codes that facilitate the editing of diagram components to prepare training samples. In the absence of such codes within a given dataset, image decomposition techniques, such as image derendering\cite{wu2017neural}~\cite{xu2021image2emmet} or vectorisation~\cite{egiazarian2020deep}~\cite{reddy2021im2vec}, are required to convert diagram images into an editable format before the preparation of the training samples. However, the success of the method is contingent on the accuracy of the conversion process, which may introduce additional challenges and variability in performance.

Furthermore, a critical component of our method, the DO loss, assumes that the embedding vectors are approximately situated within Euclidean space. In practice, however, the actual embedding spaces may not adhere to this assumption, which can limit the generalisability of the approach.
Further work is required to explore more flexible loss functions that account for non-Euclidean embedding spaces to enhancing the applicability and effectiveness of our approach in diverse diagrammatic contexts.

\section{Conclusion}
In this paper, we have presented a novel training methodology tailored to address the distinctive challenges posed by diagrammatic images in VLMs. By concentrating on the unique structural and semantic attributes of diagrammatic content, we have developed an approach that significantly improves a model's capacity to understand and interpret the intricate relationships between visual elements and their associated textual annotations. Our method incorporates hard sample mining for contrastive learning and utilisation of two specialised loss functions that capitalise on the inherent structural properties of diagrams, demonstrating that the performance of our method surpasses ones of other CLIP finetuning methods on multiple vision and language tasks.

In conclusion, the proposed method represents a significant advancement toward more effective and efficient VLMs capable of navigating the complexities of diagrammatic content. Future research will focus on refining training objectives, exploring additional forms of hard sample synthesis, and extending the framework to accommodate other structured visual data, such as charts, graphs, and a variety of diagram types.
We envision that our work will lay the groundwork for the development of more robust multimodal systems over complex visual and textual information across diverse applications.

{
    \small
    \bibliographystyle{ieeenat_fullname}
    \bibliography{references}

\begin{thebibliography}{31}
\providecommand{\natexlab}[1]{#1}
\providecommand{\url}[1]{\texttt{#1}}
\expandafter\ifx\csname urlstyle\endcsname\relax
  \providecommand{\doi}[1]{doi: #1}\else
  \providecommand{\doi}{doi: \begingroup \urlstyle{rm}\Url}\fi

\bibitem[Chen et~al.(2020{\natexlab{a}})Chen, Kornblith, Norouzi, and Hinton]{chen2020simple}
Ting Chen, Simon Kornblith, Mohammad Norouzi, and Geoffrey Hinton.
\newblock A simple framework for contrastive learning of visual representations.
\newblock In \emph{Intl. Conf. on Machine Learning}, 2020{\natexlab{a}}.

\bibitem[Chen et~al.(2020{\natexlab{b}})Chen, Li, Yu, El~Kholy, Ahmed, Gan, Cheng, and Liu]{chen2020uniter}
Yen-Chun Chen, Linjie Li, Licheng Yu, Ahmed El~Kholy, Faisal Ahmed, Zhe Gan, Yu Cheng, and Jingjing Liu.
\newblock Uniter: Universal image-text representation learning.
\newblock In \emph{European Conf. on Computer Vision}, 2020{\natexlab{b}}.

\bibitem[Devlin et~al.(2019)Devlin, Chang, Lee, and Toutanova]{devlin2019bert}
Jacob Devlin, Ming-Wei Chang, Kenton Lee, and Kristina Toutanova.
\newblock Bert: Pre-training of deep bidirectional transformers for language understanding.
\newblock In \emph{2019 Conf. of the North American Chapter of the Association for Computational Linguistics}, 2019.

\bibitem[Egiazarian et~al.(2020)Egiazarian, Voynov, Artemov, Volkhonskiy, Safin, Taktasheva, Zorin, and Burnaev]{egiazarian2020deep}
Vage Egiazarian, Oleg Voynov, Alexey Artemov, Denis Volkhonskiy, Aleksandr Safin, Maria Taktasheva, Denis Zorin, and Evgeny Burnaev.
\newblock Deep vectorization of technical drawings.
\newblock In \emph{European Conf. on Computer Vision}, 2020.

\bibitem[Faghri et~al.(2018)Faghri, Fleet, Kiros, and Fidler]{faghri2018vse++}
Fartash Faghri, David~J Fleet, Jamie~Ryan Kiros, and Sanja Fidler.
\newblock Vse++: Improving visual-semantic embeddings with hard negatives.
\newblock In \emph{British Machine Vision Conf.}, 2018.

\bibitem[He et~al.(2020)He, Fan, Wu, Xie, and Girshick]{he2020momentum}
Kaiming He, Haoqi Fan, Yuxin Wu, Saining Xie, and Ross Girshick.
\newblock Momentum contrast for unsupervised visual representation learning.
\newblock In \emph{IEEE/CVF Conf. on Computer Vision and Pattern Recognition}, 2020.

\bibitem[Hu et~al.(2022)Hu, Wallis, Allen-Zhu, Li, Wang, Wang, Chen, et~al.]{hu2022lora}
Edward~J Hu, Phillip Wallis, Zeyuan Allen-Zhu, Yuanzhi Li, Shean Wang, Lu Wang, Weizhu Chen, et~al.
\newblock Lora: Low-rank adaptation of large language models.
\newblock In \emph{10th Intl. Conf. on Learning Representations}, 2022.

\bibitem[Kim et~al.(2022)Kim, Hong, Yim, Nam, Park, Yim, Hwang, Yun, Han, and Park]{kim2022ocr}
Geewook Kim, Teakgyu Hong, Moonbin Yim, JeongYeon Nam, Jinyoung Park, Jinyeong Yim, Wonseok Hwang, Sangdoo Yun, Dongyoon Han, and Seunghyun Park.
\newblock Ocr-free document understanding transformer.
\newblock In \emph{European Conf. on Computer Vision}, 2022.

\bibitem[Li et~al.(2019)Li, Yatskar, Yin, Hsieh, and Chang]{harold2019visualbert}
Liunian~Harold Li, Mark Yatskar, Da Yin, Cho{-}Jui Hsieh, and Kai{-}Wei Chang.
\newblock Visualbert: {A} simple and performant baseline for vision and language.
\newblock \emph{CoRR}, abs/1908.03557, 2019.

\bibitem[Liu(2023)]{llava:v1.6-mistral-7b}
Haotian Liu.
\newblock llava-v1.6-mistral-7b.
\newblock \url{https://huggingface.co/liuhaotian/llava-v1.6-mistral-7b}, 2023.

\bibitem[Liu et~al.(2023)Liu, Li, Wu, and Lee]{liu2024visual}
Haotian Liu, Chunyuan Li, Qingyang Wu, and Yong~Jae Lee.
\newblock Visual instruction tuning.
\newblock In \emph{Advances in Neural Information Processing Systems 36}, 2023.

\bibitem[Liu et~al.(2024)Liu, Li, Li, and Lee]{liu2024improved}
Haotian Liu, Chunyuan Li, Yuheng Li, and Yong~Jae Lee.
\newblock Improved baselines with visual instruction tuning.
\newblock In \emph{IEEE/CVF Conf. on Computer Vision and Pattern Recognition}, 2024.

\bibitem[Lu et~al.(2019)Lu, Batra, Parikh, and Lee]{lu2019vilbert}
Jiasen Lu, Dhruv Batra, Devi Parikh, and Stefan Lee.
\newblock Vilbert: Pretraining task-agnostic visiolinguistic representations for vision-and-language tasks.
\newblock In \emph{Advances in Neural Information Processing Systems 32}, 2019.

\bibitem[Masry et~al.(2023)Masry, Kavehzadeh, Long, Hoque, and Joty]{masry2023unichart}
Ahmed Masry, Parsa Kavehzadeh, Do~Xuan Long, Enamul Hoque, and Shafiq Joty.
\newblock Unichart: A universal vision-language pretrained model for chart comprehension and reasoning.
\newblock In \emph{Conf. on Empirical Methods in Natural Language Processing}, 2023.

\bibitem[{Mermaid Chart}()]{std:mermaid}
{Mermaid Chart}.
\newblock {Mermaid Diagramming and charting tool}.
\newblock \url{https://mermaid.js.org}.

\bibitem[mermaid js(2024)]{code:mermaid-cli}
mermaid js.
\newblock {mermaid-cli}.
\newblock \url{https://github.com/mermaid-js/mermaid-cli/tree/6e593e3b81aacaf7984151b96d7644daa58daa41}, 2024.
\newblock Version 10.9.1.

\bibitem[OpenAI(2021)]{clip:vit-large-patch14}
OpenAI.
\newblock clip-vit-large-patch14-336.
\newblock \url{https://huggingface.co/openai/clip-vit-large-patch14-336}, 2021.

\bibitem[Paiss et~al.(2022)Paiss, Chefer, and Wolf]{paiss2022no}
Roni Paiss, Hila Chefer, and Lior Wolf.
\newblock No token left behind: Explainability-aided image classification and generation.
\newblock In \emph{European Conf. on Computer Vision}, 2022.

\bibitem[Paiss et~al.(2023)Paiss, Ephrat, Tov, Zada, Mosseri, Irani, and Dekel]{paiss2023teaching}
Roni Paiss, Ariel Ephrat, Omer Tov, Shiran Zada, Inbar Mosseri, Michal Irani, and Tali Dekel.
\newblock Teaching clip to count to ten.
\newblock In \emph{IEEE/CVF Intl. Conf. on Computer Vision}, 2023.

\bibitem[Patel et~al.(2024)Patel, kusumba, Cheng, Kim, Gokhale, Baral, and Yang]{patel2024tripletclip}
Maitreya Patel, Naga Sai~Abhiram kusumba, Sheng Cheng, Changhoon Kim, Tejas Gokhale, Chitta Baral, and Yezhou Yang.
\newblock Triplet{CLIP}: Improving compositional reasoning of {CLIP} via synthetic vision-language negatives.
\newblock In \emph{Advances in Neural Information Processing Systems 37}, 2024.

\bibitem[Radford et~al.(2021)Radford, Kim, Hallacy, Ramesh, Goh, Agarwal, Sastry, Askell, Mishkin, Clark, et~al.]{radford2021learning}
Alec Radford, Jong~Wook Kim, Chris Hallacy, Aditya Ramesh, Gabriel Goh, Sandhini Agarwal, Girish Sastry, Amanda Askell, Pamela Mishkin, Jack Clark, et~al.
\newblock Learning transferable visual models from natural language supervision.
\newblock In \emph{Intl. Conf. on Machine Learning}, 2021.

\bibitem[Reddy et~al.(2021)Reddy, Gharbi, Lukac, and Mitra]{reddy2021im2vec}
Pradyumna Reddy, Michael Gharbi, Michal Lukac, and Niloy~J Mitra.
\newblock Im2vec: Synthesizing vector graphics without vector supervision.
\newblock In \emph{IEEE/CVF Conf. on Computer Vision and Pattern Recognition}, 2021.

\bibitem[Ren et~al.(2017)Ren, He, Girshick, and Sun]{ren2017faster}
Shaoqing Ren, Kaiming He, Ross Girshick, and Jian Sun.
\newblock Faster r-cnn: Towards real-time object detection with region proposal networks.
\newblock \emph{IEEE Trans. on Pattern Analysis and Machine Intelligence}, 39\penalty0 (6):\penalty0 1137--1149, 2017.

\bibitem[Schuhmann et~al.(2022)Schuhmann, Beaumont, Vencu, Gordon, Wightman, Cherti, Coombes, Katta, Mullis, Wortsman, Schramowski, Kundurthy, Crowson, Schmidt, Kaczmarczyk, and Jitsev]{schuhmann2022laionb}
Christoph Schuhmann, Romain Beaumont, Richard Vencu, Cade~W Gordon, Ross Wightman, Mehdi Cherti, Theo Coombes, Aarush Katta, Clayton Mullis, Mitchell Wortsman, Patrick Schramowski, Srivatsa~R Kundurthy, Katherine Crowson, Ludwig Schmidt, Robert Kaczmarczyk, and Jenia Jitsev.
\newblock {LAION}-5b: An open large-scale dataset for training next generation image-text models.
\newblock In \emph{Advances in Neural Information Processing Systems 35}, 2022.

\bibitem[Singh et~al.(2024)Singh, Chaurasia, Varun, Pandya, Gupta, Gupta, and Roth]{singh2024flowvqa}
Shubhankar Singh, Purvi Chaurasia, Yerram Varun, Pranshu Pandya, Vatsal Gupta, Vivek Gupta, and Dan Roth.
\newblock {F}low{VQA}: Mapping multimodal logic in visual question answering with flowcharts.
\newblock In \emph{Findings of the Association for Computational Linguistics}, 2024.

\bibitem[van~den Oord et~al.(2018)van~den Oord, Li, and Vinyals]{oord2018representation}
A{\"{a}}ron van~den Oord, Yazhe Li, and Oriol Vinyals.
\newblock Representation learning with contrastive predictive coding.
\newblock \emph{CoRR abs/1807.03748}, 2018.

\bibitem[W3C()]{std:svg}
W3C.
\newblock {Scalable Vector Graphics}.
\newblock \url{https://www.w3.org/Graphics/SVG/}.

\bibitem[Wu et~al.(2017)Wu, Tenenbaum, and Kohli]{wu2017neural}
Jiajun Wu, Joshua~B Tenenbaum, and Pushmeet Kohli.
\newblock Neural scene de-rendering.
\newblock In \emph{IEEE/CVF Conf. Computer Vision and Pattern Recognition}, 2017.

\bibitem[Xu et~al.(2021)Xu, Bo, Sun, Li, Jiang, and Zhou]{xu2021image2emmet}
Yong Xu, Lili Bo, Xiaobing Sun, Bin Li, Jing Jiang, and Wei Zhou.
\newblock image2emmet: Automatic code generation from web user interface image.
\newblock \emph{Journal of Software: Evolution and Process}, 33\penalty0 (8):\penalty0 e2369, 2021.

\bibitem[Zhang et~al.(2025)Zhang, Wei, Jiang, Guo, Zhang, Tong, Liu, Zhou, Zhang, Gao, and Li]{zhang2025mavis}
Renrui Zhang, Xinyu Wei, Dongzhi Jiang, Ziyu Guo, Yichi Zhang, Chengzhuo Tong, Jiaming Liu, Aojun Zhou, Shanghang Zhang, Peng Gao, and Hongsheng Li.
\newblock {MAVIS}: Mathematical visual instruction tuning with an automatic data engine.
\newblock In \emph{13th Intl. Conf. on Learning Representations}, 2025.

\bibitem[Zhang et~al.(2020)Zhang, Kishore, Wu, Weinberger, and Artzi]{zhang2020bertscore}
Tianyi Zhang, Varsha Kishore, Felix Wu, Kilian~Q Weinberger, and Yoav Artzi.
\newblock Bertscore: Evaluating text generation with bert.
\newblock In \emph{8th Intl. Conf. on Learning Representations}, 2020.

\end{thebibliography}
}

\end{document}